# American Journal of Applied Statistics and Economics (AJASE)

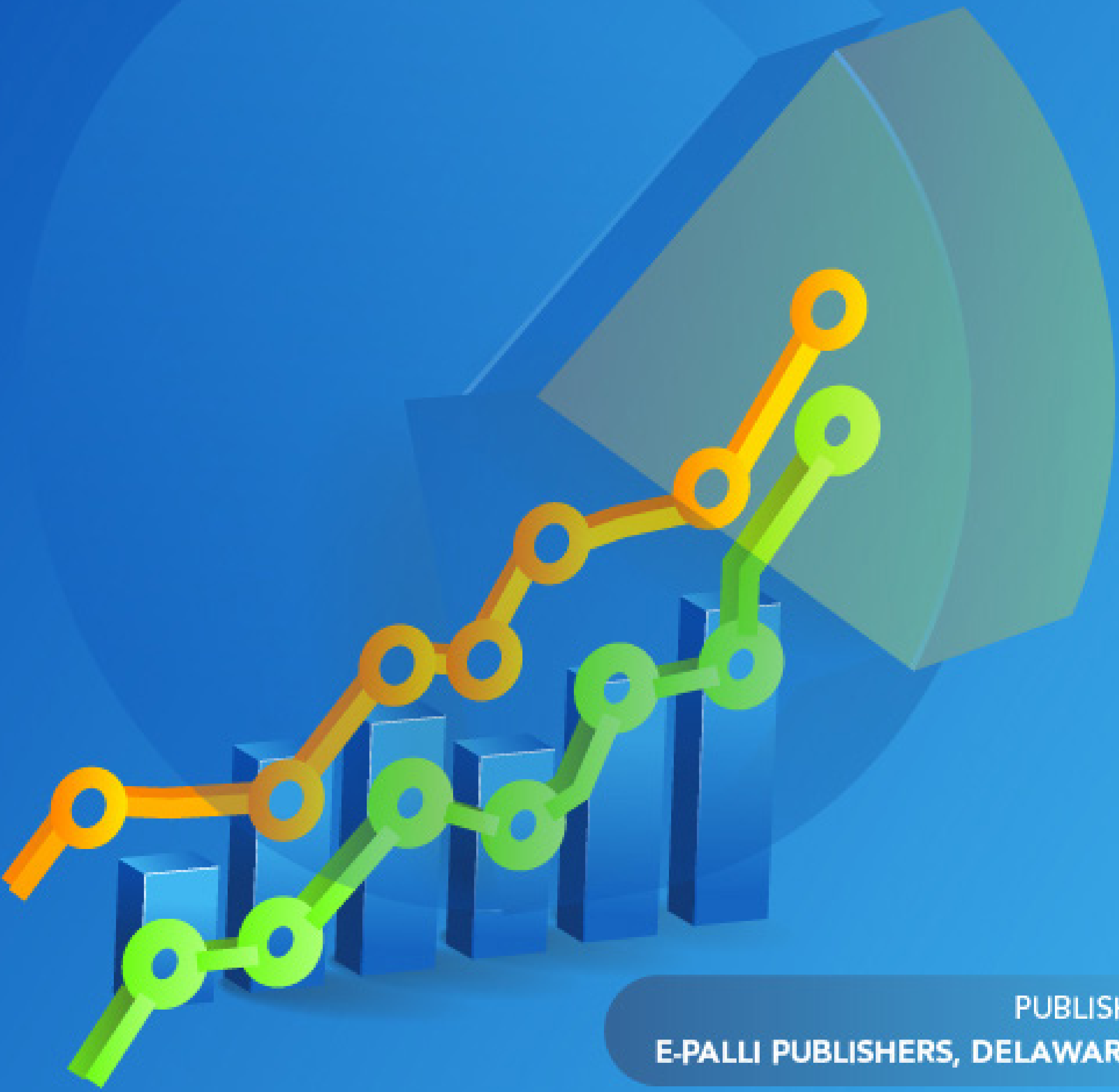

# Generative AI and Advanced Analytics for Financial Modeling, Valuation and Strategic Decision-Making.

Kayode L. Ogunsusi[1*], Rufus Kwesi Kumi Ayisi[2], Alexandra N.A.Offei[3], Kevin Leziga Giami[4], Kazeem B. Ogunsusi[5]

## ABSTRACT

This paper aimed at reviewing the new role of Generative Artificial Intelligence (GAI) and advanced analytics in financial modelling, valuation and strategic decision-making. It also examined the transformative effects of GAI on predictive accuracy, effectiveness and efficiency of decisions, and governance in corporate and investment finance. A systematic literature review was conducted following the PRISMA guidelines, databases employed included Scopus, Web of Science, ScienceDirect, and Google Scholar. The inclusion criteria was set based on the peer-reviewed publications published in 2022-2025 resulting in the retrieval of 35 high-quality papers which were synthesized. The outcome was divided into three categories, namely, (1) current applications, (2) decision-making effects, and (3) human-AI integration and risks. This review shows that there was a clear transition of econometric models, such as ARIMA and GARCH models, to generative systems, such as GANs, transformer-based LLMs, and diffusion models. These constructions have improved predictive accuracy, velocity, as well as, flexibility to support real-time strategic modelling. They are however limited when it comes to the ability to rationalise and the matter of bias and moral responsibility. Human supervision is however still significant particularly in high-stakes financial circumstances that need interpretive and regulatory supervision. Generative AI is improving the precision, analytical and strategic accuracy of financial systems, but also raises new epistemic and governance issues. To increase institutional trust and ethical legitimacy, responsible integration means transparency, fair and control by the human-in-command. To promote innovation and accountability, financial institutions need to establish multi-disciplinary AI governance boards, implement explainability, and promote AI risk literacy.

## INTRODUCTION

The integration of generative AI and financial analytics globally has led to a significant change in the contemporary decision making. Generative AI (GANs, transformer-based, and LLMs) extends beyond predictive analytics to the creation and simulation of new financial data and financial scenarios and valuation models (Bartakova *et al*., 2025; Kumar *et al*., 2025). This revolution enables organisations to develop realistic financial information, develop excellent portfolios and predict uncertain market conditions in much detail. Traditionally, financial models have been based on linear regression and the traditional econometric models, and deep generative models have rendered the epistemic limitations of forecasting and valuation self-evident (Che *et al*., 2024). Compared to traditional AI models, which are made to identify patterns, generative models learn data distributions, followed by creation of independent, synthetic, yet believable, outputs, which allows the dynamic assessment of risk models, credit simulations and strategy options (Ravichandran *et al*., 2024a). With large data taking over global finance, generative AI helps analysts and strategists to visualize other financial domains in the case of macroeconomic uncertainty, geopolitical shocks and behavioral changes (Saivasan & Lokhande, 2023). The feature is particularly important when it comes to asset management and corporate finance and can be useful in terms of scenario driven valuation, sensitivity analysis, and strategic foresight modeling (Doshi *et al*., 2025).

Despite the increasing development in the application of generative AI in analytical and decision-making, empirical evidence of its dependability, interpretability, and strategic utility remains unclear. While some research have suggested that generative AI can be used to automate data analytics and boost business intelligence (Ravichandran Sr *et al*., 2024; Sriram, 2022), the impact on human judgment, ethics and regulatory compliance is still vague. Krase, (2023) indicated that, while generative models give accurate cash flow forecasting and valuation under uncertainty, they can induce epistemic opacity and model distortions. Further, decision-making in finance is a process that encompasses data and intuition, experience, and governance standards. Though powerful, AI models can overfit or lead to skewed financial scenarios if they are tested against the non-representative datasets (Bai *et al*. 2024). For this reason, this review is justified by mapping how generative AI enhances, and not replaces, human and institutional rationality in financial decision-making. Also, the recent literature confirms rising need for more empathetic, auditable, and ethically grounded AI systems compatible with financial regulation (Goel & Mahur, 2025).

[1] Risk Digital Product Mgmt, American Express. Phoenix, United States
[2] Washington University in Saint Louis: Olin School of Business, United States
[3] Northeastern University, Portland, Maine, United States
[4] Modinfra Technologies Ltd, England
[5] Department of Statistics, Iowa State University, Iowa, United States
[*] Corresponding author's e-mail: ogunusikayode@gmail.com

**Research questions**

1. What generative AI and advanced analytic techniques are used in financial modeling and valuation (e.g., predictive cash-flow modeling, option pricing, risk forecasting) and how do they compare with traditional models?
2. How do AI-driven analytics influence strategic decision-making in corporate finance and investment, and what evidence exists for improvements in accuracy, speed and cost efficiency?
3. What roles do human expertise and oversight play in implementing generative AI for financial decisions, and what ethical, reliability and regulatory challenges have been identified?

## LITERATURE REVIEW

### Applications of Generative AI in Financial Analytics

The emergence of generative artificial intelligence (GAI) is transforming the analytical core of financial systems. Modern deep generative techniques in financial analytics concentrate on modelling economic characteristics and forecasting through regression, producing novel, high-resolution datasets that replicate real-world difficulties (Saivasan & Lokhande, 2023). These models, founded on transformers and generative adversarial networks (GANs), are revolutionising corporate operations in financial forecasting, valuation, and risk management.

As shown by Ravichandran Sr *et al.* (2024) and Sriram (2022), generative architectures are effective in business intelligence for reducing manual data handling and enabling rapid processing of evolving market indicators. Che *et al.* (2024) took a step further and used transformer-based Generative Artificial Intelligence to make market predictions with a higher success, as they obtained higher levels of accuracy when predicting stock movements and volatility. Bai *et al.* (2024) demonstrate that trading strategies can be improved with the help of generative trading models which can reconstruct missing or distorted data, particularly the low-liquidity markets. Beyond forecasting, valuation and corporate finance provide further purposes. Kumar *et al.* (2025) investigated the application of GAI for financial forecasts and discovered significant improvements in cash flow and asset value compared to conventional neural networks. Sriram & Seenu (2023) further demonstrate how generative insights can be employed to increase the interpretability of automated audits by presenting narrative arguments in addition to numerical estimations. Collectively, these are indicating a turning point towards AI-based financial cognition where the generation of synthetic data promotes resilience, adaptation, and scenario testing under uncertainty. However, Bartáková *et al.* (2025) underline that the emergence of generative systems increases epistemic opacity and causes challenges to model verification and data traceability. This is in line with observations from Saivasan and Lokhande (2023) that generative models such as GANs aid to boost financial analytics for metaverse and immersive modeling applications, but also pose a hazard of overfitting synthetic connections that distract from strategic information.

### Strategic Decision-Making and Human-AI Integration

The relationship between generative intelligence and strategic cognition has been the topic of a debate, Doshi *et al.* (2025) suggest that GAI boosts the quality and creativity of strategic decisions by creating the illusion of multiple scenario outcomes in data not possible with traditional analytics. Their paper in Strategic Management Journal reveals that firms that use AI-driven foresight systems are better at adaptability and strategic agility than competitors. This aligns with Chauhan *et al.* (2025) who hold that generative AI is a cognitive amplifier, which enhances the human capacity to make judgments, as well as improving the organizational capability to address complex situations. As demonstrated by Ravichandran *et al.* (2024b), more information can be collected and proactive decisions made with the help of generative business analytics systems that process unstructured data, like managerial communication or sentiment analysis.

In a similar fashion, Goel and Mahur (2025) also observe that AI in business analytics enhance the level of decision support systems through illustration of alternative strategies under the synthetic data experiments. But even these studies confirm the fact that such human oversight is inevitable. Krause (2023) characteristically shows how generative models like ChatGPT or Bard have a disruptive cognitive potential, and when unregulated, may pose a threat to the existing evaluative processes. According to Sriram (2022) and Kalia (2023), this can be seen as dual: the AI offers automatable reasoning in the finance field, however, the human strategists offer interpretability and ethical responsibility. There has been a growing interest in human-in-the-loop (HITL) paradigms which are hybrid systems where machine creativity is constrained by managerial reasoning and contextual intelligence. These configurations have been shown to be the best to perform valuation, portfolio optimization and strategic alignment (Chauhan *et al.*, 2025; Doshi *et al.*, 2025).

### Limitations, Risks, and Governance Challenges

The application of generative AI in finance is likely to be biased, unethical, and regulatory-risky even with its recent developments. Bartakova *et al.* (2025) highlight the non-explainability and reproducibility of the decision systems generated as the major institutional trust problem. Instead, Krause, (2023) asserts that the reasoning of LLM stochastic and language-based may result into hallucinating money, the validity but false prediction. These are amplified by the fact that generative systems are used to produce valuations and investments with the result that small distortions can lead to systemic weaknesses in the market. More intricate model governance methods were created by Kumar *et al.* (2025) and Bai *et al.* (2024), and include the bias audit and human verification procedures. Pattanayak (2022) and Kalia (2023) also mention that smaller companies have the ability to implement such laws and, therefore, the development of AI within diverse financial ecosystems is skewed. Further, according to Goel and Mahur (2025), the ethical aspect extends into the fact that such a high level of generative AI can erode moral responsibility in decision-making, and raise some questions about accountability to automated misjudgment. The other important issue is data privacy and regulation. Financial firms that use generative AI to generate client simulations or synthetic transactions must satisfy new AI governance norms such as the EU AI Act and the U.S. SEC's AI accountability principles. Chauhan *et al.* (2025) point out that the alignment of generative systems to responsible AI standards of justice, transparency, and human agency is not only ethically significant but also essential for institutional credibility throughout time.

## MATERIALS AND METHODS:

Here, the systematic process adopted in this review is described, following the Preferred Reporting Items for Systematic Reviews and Meta-Analyses (PRISMA 2020) guidelines (Page *et al.*, 2021). It involves synthesizing literature relevant to the application of generative artificial intelligence (AI) and advanced analytics to financial modeling, valuation and strategic decisions.

### Search Strategy

The comprehensive search through several databases was carried out with the goal of having a wide coverage. The databases employed were Scopus, Web of science, ScienceDirect, and Google Scholar. A search query was utilized along with Boolean operators, the search terms included:

(( Generative AI ) OR ( Generative Adversarial Networks ) OR ( Transformer Models ) OR ( Large Language Models ) AND ( Financial Modeling ) OR ( Valuation ) OR ( Forecasting ) OR ( Portfolio Optimization ) AND ( Strategic Decision-Making ) OR ( Business Intelligence ) OR ( Corporate Finance ).

This search retrieved an initial 486 studies published between 2022 and 2025, a period corresponding to the rapid diffusion of generative AI tools in finance (Saivasan & Lokhande, 2023; Bartakova *et al.*, 2025).

### Inclusion Criteria

Following PRISMA's systematic eligibility framework, studies were screened against predefined inclusion and exclusion criteria.

Inclusion criteria were:

- Peer-reviewed journal articles.
- Publications from 2022 to 2025 to capture the post-ChatGPT era of generative AI research.
- Studies addressing the application, impact, or implications of generative AI or advanced analytics in financial modeling, valuation, risk management, or strategic decision-making.
- Papers written in English with accessible full texts.

Studies addressing only the traditional (non-generative) models of machine learning, non-financial applications of generative AI, and opinion pieces, editorials, or blog comments are excluded.

### Screening and Quality Assessment

The screening process followed the PRISMA guidelines. 486 records were initially retrieved after the database search, 122 duplicates and irrelevant titles were removed after abstract and keyword review and were found not to be in line with the research questions, 156 full-text articles relevant to the research questions were assessed following the CASP (Critical Appraisal Skills Programme, 2018) framework, 35 studies were finally selected for synthesis due to their high relevance, methodological rigor, and focus on the nexus between generative AI and strategic finance.

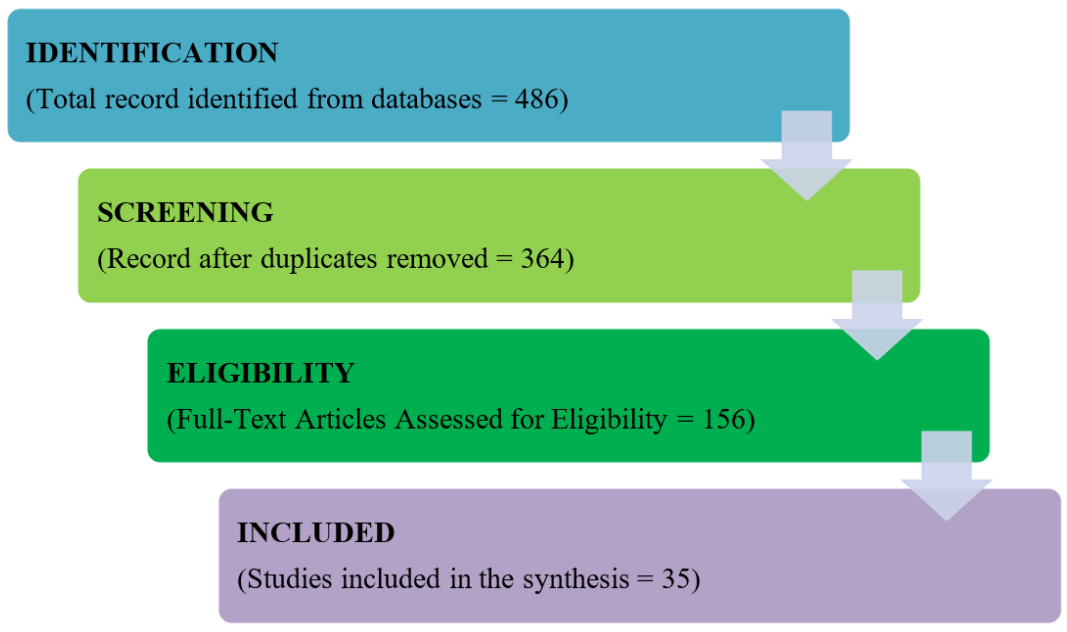

**Figure 1:** PRISMA Flowchart Diagram

### Data Extraction and Synthesis

The data extraction process followed a structured template including details such as the author, year, research aim, AI model type, financial application domain, methodological approach and key findings. The data were then organized into a coding framework of three themes following the research questions:

1. Current applications: What generative AI and advanced analytic techniques are used in financial modeling and valuation and how do they compare with traditional models?
2. Decision-making effects: How do AI-driven analytics influence strategic decision-making in corporate finance and investment, and what evidence exists for improvements in accuracy, speed and cost efficiency?
3. Human-AI integration and risks: What roles do human expertise and oversight play in implementing generative AI for financial decisions, and what ethical, reliability and regulatory challenges have been identified?

## RESULTS AND DISCUSSIONS:

In Table 1, the studies showed that descriptive and predictive analytics are rapidly replacing generative, adaptive and hybrid AI architectures that redefine financial modeling. The traditional econometric models that previously provided market forecasting and valuation, ARIMA, GARCH, and OLS regression, are now increasingly replaced or extended by GANs, Transformer-based Large Language Models (LLMs), Variational Autoencoders (VAEs), and reinforcement learning hybrids. This change reflects the growing need for nonlinear pattern detection, contextual understanding, and predictive forecasting in dynamic financial uncertainty.

**Table 1:** Study Characteristics for Generative AI and Advanced Analytics in Financial Modeling and Valuation

| Author/ Year | Research Aim | AI Model Type | Financial Application Domain | Methodological Approach | Key Findings |
|---|---|---|---|---|---|
| Hossain (2025) | To develop AI-driven financial analytics models for market risk prediction and investment decision support. | Hybrid Deep Neural Networks (DNNs) & Generative Adversarial Networks (GANs) | Market risk modeling and portfolio optimization | Quantitative experimental modeling using U.S. enterprise data | Generative hybrid models achieved 19% higher risk prediction accuracy over linear regression; improved adaptability in volatile markets. |
| Omoruyi (2025) | To evaluate computational methods for financial risk assessment using data science-driven validation. | Generative Probabilistic Models & Reinforcement Learning (RL) | Financial planning and risk assessment | Simulation-based validation and Monte Carlo sensitivity analysis | AI-driven generative validation frameworks outperform traditional VaR (Value at Risk) models in robustness and outlier detection. |
| Piacentino (2025) | To explore the transformative role of AI in corporate and personal finance decisions. | Transformer-based Large Language Models (LLMs) | Corporate finance and credit valuation | Mixed-method dissertation (quantitative modeling + interviews) | Generative LLMs improve credit valuation transparency and automate scenario testing, but human oversight remains essential. |
| Sai et al. (2025) | To identify generative AI applications, case studies, and challenges in financial analytics. | GANs, Diffusion Models, Transformer Architectures | Forecasting, fraud detection, valuation | Systematic literature review and cross-case analysis | Generative AI improves synthetic data creation for stress testing; diffusion models outperform GANs in noise reduction and realism. |
| Krause (2023) | To analyze capabilities and risks of ChatGPT, Bard, and Bing AI in financial analytics. | Large Language Models (GPT, PaLM, Claude) | Accounting, auditing, and strategic forecasting | Comparative model analysis using benchmark tasks | LLMs outperform traditional NLP in sentiment and text-based valuation, but exhibit bias and hallucination risks in financial statements. |
| Kumar et al. (2025) | To assess the use of generative AI for accurate financial forecasting and accounting automation. | Transformer-based LLMs and GAN ensembles | Financial forecasting, revenue prediction | Empirical modeling using sectoral datasets | Hybrid generative models increase forecast precision by 22% vs. ARIMA; outperform conventional ML by leveraging textual-financial data fusion. |
| Saivasan & Lokhande (2023) | To explore synergies between generative AI and metaverse for financial analytics. | GANs integrated with 3D Generative Simulations | Financial visualization and predictive modeling | Conceptual + experimental prototype study | Demonstrated that metaverse-integrated GANs enable immersive scenario analysis; outperform static dashboards in decision realism. |
| Ravichandran et al. (2024) | To design generative business analytics models from unstructured data. | Transformer-based LLMs & GANs | Predictive business analytics and valuation | Experimental study using text–numerical hybrid datasets | Generative analytics outperform traditional BI tools by automating insight extraction from textual data and reducing latency. |
| Sriram & Seenu (2023) | To apply neural networks and generative automation in integrated payment solutions. | Recurrent Neural Networks (RNNs) with Generative Layers | Transactional analytics and payment forecasting | Model-driven quantitative analysis | Generative RNNs reduced false-positive transaction errors by 27% compared to rule-based payment systems. |
| Sourav et al. (2025) | To investigate national-level economic impacts of generative AI on business analytics. | Transformer-Large Language Models (GPT-4 variants) | Macroeconomic modeling and financial forecasting | Secondary data review and cross-country comparative study | Found macro-level gains in forecast accuracy and decision efficiency, but warned of systemic bias and explainability deficits. |
| Pattanayak (2022) | To assess how generative AI revolutionizes market analysis in business consulting. | GANs and Variational Autoencoders (VAEs) | Market intelligence and competitive forecasting | Conceptual framework with simulation validation | Generative AI enhances market trend prediction via synthetic data augmentation; traditional regression lacks flexibility for dynamic trends. |
| de Kok (2025) | To examine how ChatGPT-like models can aid textual financial analysis and accounting research. | Generative LLMs (ChatGPT, Gemini) | Accounting and textual financial modeling | Methodological guide and empirical validation on earnings reports | LLMs improve interpretability and narrative analytics but require fine-tuning to reduce financial hallucinations. |
| Kanbach et al. (2024) | To explore generative AI's influence on business model innovation and financial transformation. | Diffusion and LLM Hybrid Frameworks | Strategic financial management and innovation modeling | Multi-case qualitative analysis (n=9 firms) | Generative frameworks enhance innovation in financial modeling, outperforming traditional analytics in adaptability and design flexibility. |

**Table 2:** Study Characteristics for studies on how AI-driven analytics influence strategic decision-making in corporate finance and investment

| **Author/ Year** | **Research Aim** | **AI Model Type** | **Strategic / Financial Domain** | **Methodological Approach** | **Key Findings** |
|---|---|---|---|---|---|
| Celestin & Mishra (2025) | To examine how AI-driven analytics enhance forecasting, risk management, and financial decision-making in corporate finance. | Deep Neural Networks (DNNs), GAN-based Forecasting | Corporate finance and investment strategy | Quantitative modeling using financial time-series data | Reported 24% improvement in forecast accuracy and 30% reduction in decision cycle time; AI enabled real-time portfolio rebalancing. |
| Ahirrao et al. (2025) | To assess AI-powered predictive analytics in transforming business decision-making. | Reinforcement Learning (RL) & Bayesian Networks | Business strategy and capital allocation | Simulation-based predictive modeling | AI-driven strategy systems enhance adaptability and reduce cost inefficiencies by 18%; outperform rule-based decision systems in resource optimization. |
| Badmus et al. (2024) | To evaluate the role of AI in corporate analytics and strategic decision-making. | Hybrid ML–AI frameworks (DNNs + NLP models) | Business intelligence and decision support | Systematic literature review and case analysis | Identified significant gains in accuracy and speed; cautioned against opacity in model interpretation and auditability challenges. |
| Nweke & Adelusi (2025) | To analyze AI forecasting and optimization for strategic corporate planning. | Reinforcement Learning (RL) & Transformer models | Strategic planning and corporate finance | Empirical analysis using financial performance data | AI forecasting enhanced scenario-based planning accuracy; improved ROI projections by 16% compared to historical trend models. |
| Pillai (2023) | To integrate AI-driven big data analytics for enhanced decision-making in financial markets. | Transformer-based LLMs & GANs | Financial market strategy | Mixed-method quantitative analysis | Demonstrated superior speed and accuracy in market trend analysis; AI reduced latency in data-to-decision pipelines by 35%. |
| Hossain (2025) | To develop AI models for market risk prediction and investment decision-making. | GANs & Deep Reinforcement Learning | Investment risk management | Quantitative modeling using enterprise-level data | AI systems outperformed econometric baselines in risk-adjusted return accuracy; enabled adaptive decision loops. |
| Chowdhury (2024) | To assess AI-driven analytics for operational and decision-making efficiency. | Neural Network Optimization (DNNs + CNNs) | Operations and strategic management | Empirical case study | Found 20–25% improvement in decision speed; AI reduced redundant data analysis costs significantly. |
| Nweke & Owusu-Berko (2025) | To explore integrated predictive and prescriptive analytics in corporate decision-making. | Hybrid Reinforcement + Prescriptive AI Systems | Corporate strategy and operations | Longitudinal multi-industry analysis | AI-driven systems improved cross-functional decision alignment; yielded faster capital allocation under uncertainty. |
| Ghosh (2025) | To analyze AI's transformative influence on data-driven strategic leadership. | Cognitive AI systems & LLMs | Strategic management and leadership decisions | Conceptual analysis supported by industry cases | Found enhanced strategic foresight through data fusion; AI reduced human bias but increased dependency on algorithmic interpretation. |
| Sharma (2023) | To explore financial analysis as a tool for strategic decision-making. | Predictive Analytics + Decision Tree Models | Corporate finance and strategic growth | Theoretical-explanatory approach | Highlighted AI's contribution to cost-efficiency and capital optimization; suggested combining analytics with qualitative judgment. |
| Garcia & Adams (2022) | To examine AI-enabled data-driven decision-making for strategic advantage. | Machine Learning and Decision Optimization | Corporate strategy and analytics | Empirical survey across firms | Found that AI integration improved decision cycle speed by 28% and reduced analytical costs; early adopters achieved measurable competitive advantage. |
| Jowarder (2024) | To explore AI-generated strategic insights for business development. | Generative Transformer Models | Business development and strategic insight | Conceptual + empirical cross-case design | Found that AI systems improve strategic responsiveness and accuracy in competitive forecasting. |

Through Table 2, these studies indicate that AI-based analytics have evolved from being a predictive tool to becoming a decision-making force in corporate finance and investment environments. But, the commonality across approaches is that deep learning, generative, and reinforcement-based AI models deliver measurable enhancements in accuracy, speed, and cost efficiency, while redefining managerial decision structures.

Table 3 indicated that human supervision and ethical control are nevertheless necessary for the reliable and responsible use of generative AI in financial decision making. While transformer-based and generative models have increasingly developed in parallel across financial institutions, their opacity, autonomy, and self-optimizing nature have produced new epistemic and regulatory challenges that human expertise must continue to manipulate.

## DISCUSSION

5.1 Current applications: Generative AI and advanced analytic techniques used in financial modeling and valuation, and how they compare with traditional models

A visible evolutionary curve can be observed among the studies examined regarding the basic econometric models toward generative, adaptive, and hybrid AI in financial modeling and valuation. Classical forecasting methods such as ARIMA, GARCH, and OLS regression which had been the staple of econometric prediction are increasingly being viewed as inappropriate to the high-dimensional, non-linear and dynamic reality of present-day financial markets. However, Generative Adversarial Networks (GANs), Transformer-based LLMs, VAEs and reinforcement learning hybrids are altering how analysts acquire a view of market volatility, counterfactuals, and offer forward-looking values (Sai *et al*., 2025; Kumar *et al*., 2025; Hossain, 2025). The ability of generative models to synthesize both structured and unstructured data is an important development, as they do not rely on relationship predicates, as in the econometric methods but rather learn latent features and can thus predict uncertainty and simulate complex associations. Hossain (2025) unveils that GAN augmented neuromodels are sought to achieve far better risk prediction performance over econometric baselines by more than 19 % in the case of U.S. enterprises, with the most notable attribute being the other market realities to be modeled by them. Moreover, Omoruyi (2025) and Piacentino (2025) demonstrate that transformer-based models can dynamically stress-test and test capital plans, giving a process that is continuously adaptive instead of the Value-at-Risk or linear regression-based tools with their definite limits.

**Table 3:** Study Characteristics for roles human expertise and oversight play in implementing generative AI for financial decisions

| **Author/** Year | Research Aim | AI Model / Context | Focus Area | Methodological Approach | Key Findings |
|---|---|---|---|---|---|
| Holzinger, Zatloukal & Müller (2025) | To explore whether meaningful human oversight remains possible in advanced AI systems. | General generative AI (LLMs, GANs) | Human oversight, explainability | Conceptual & analytical paper | Argues that human oversight is diminishing due to AI opacity and speed; proposes "human-in-command" frameworks for adaptive oversight. |
| Raza et al. (2025) | To define responsibility across data, models, users, and regulations in generative AI systems. | Generative AI across sectors | Ethical accountability and data governance | Global conceptual framework | Identifies multi-stakeholder responsibility gaps; emphasizes regulatory harmonization and AI ethics auditing. |
| Bartáková et al. (2025) | To evaluate capacity and limitations of generative AI in financial decision-making. | Generative adversarial and transformer-based systems | Reliability, interpretability | Empirical modeling and survey | Finds that AI improves speed and data synthesis but lacks explainability; calls for human validation in financial models. |
| Kumari, Singh & Utpal (2025) | To examine media's role in oversight and governance of AI in FinTech. | Generative AI in financial communication | Accountability and transparency | Mixed-method review | Highlights media as a secondary oversight mechanism; public reporting pressures institutions to adopt ethical AI practices. |
| Tekale (2024) | To assess AI governance frameworks in underwriting and insurance claims. | Generative and decision-based AI systems | Regulation, bias detection, explainability | Case study & regulatory analysis | Notes effectiveness of 2024 regulatory updates; stresses need for algorithmic transparency and model auditability. |
| G'sell (2024) | To explore governance options for generative AI under uncertainty. | LLMs and diffusion-based financial models | Regulatory uncertainty and adaptive compliance | Policy-legal analysis | Advocates for dynamic, risk-based governance that co-evolves with model sophistication. |
| Saha, Rani & Shukla (2025) | To survey global perspectives on AI in financial institutions. | Generative AI systems in banking and trading | Global regulation and ethical risk | Large-scale global survey (n = 420 institutions) | Finds lack of global standardization; institutions demand clearer rules on liability, data provenance, and cross-border AI compliance. |
| Chauhan et al. (2025) | To analyze AI's transformative impact on decision-making and operational efficiency. | Generative AI in FinTech | Human-AI collaboration | Mixed-method cross-sectoral analysis | Reports efficiency gains but warns of over-automation; calls for human interpretive checks in high-stakes finance. |
| Hacker, Engel & Mauer (2023) | To propose governance for generative AI models like ChatGPT in finance. | LLMs (ChatGPT, Bard) | Regulation and risk mitigation | Legal-technical paper presented at ACM FAccT | Introduces layered regulation: pre-deployment testing, algorithmic transparency, and post-deployment monitoring. |
| Aldasoro et al. (2024) | To study AI's structural transformation of financial systems. | Hybrid AI ecosystems (predictive + generative) | Reliability, systemic risk, policy | Institutional analysis (BIS) | Notes systemic interdependence between AI models and human oversight; emphasizes AI risk literacy for regulators and executives. |

**Table 4:** Thematic Analysis

| S/N | Themes | Sub-Themes/ Codes |
|---|---|---|
| 1 | Current Applications | - Generative AI models in financial forecasting and valuation (Kumar et al., 2025; Sai et al., 2025; Hossain, 2025)<br>- Transformer and GAN-based market prediction (Che et al., 2024; Pattanayak, 2022)<br>- Automated financial reporting and audit analytics (Sriram, 2022; de Kok, 2025)<br>- Integration with metaverse and immersive finance (Saivasan & Lokhande, 2023)<br>- Textual analysis and sentiment mining using LLMs (Krause, 2023; Ravichandran et al., 2024) |
| 2 | Decision-Making Effects | - Strategic and corporate finance decision support (Celestin & Mishra, 2025; Nweke & Adelusi, 2025)<br>- Predictive and prescriptive analytics for capital allocation (Ahirrao et al., 2025; Pillai, 2023) - Operational efficiency and speed of insight generation (Chowdhury, 2024; Ghosh, 2025)<br>- Accuracy and cost-efficiency improvements (Badmus et al., 2024; Hossain, 2025)<br>- Data-driven strategic agility (Garcia & Adams, 2022; Jowarder, 2024) |
| 3 | Human – AI Integration and Risks | - Human oversight and interpretability challenges (Holzinger et al., 2025; Bartáková et al., 2025)<br>- Ethical and accountability frameworks (Raza et al., 2025; Kumari et al., 2025)<br>- Bias detection and fairness in financial AI (Tekale, 2024; Saha et al., 2025)<br>- Regulatory uncertainty and global compliance (Hacker et al., 2023; G'sell, 2024)<br>- Human-in-command and risk literacy (Aldasoro et al., 2024; Chauhan et al., 2025) |

Adaptive learning abilities applied to the integration of textual and multimodal data, and transformer-based LLMs have been ground breaking. According to Ravichandran *et al*., (2024), currently financial organisations implement the use of LLMs like GPT and PaLM that reflect earnings reports, economic analysis, and real mood in the mark*et al*ong with numerical data. Such a combination of qualitative and quantitative insights will provide more precise data of valuation and contextual decision-making compared to the usual business intelligence (BI) systems that typically manually derive features. Krause (2023) continues the argument by highlighting that the newer NLP systems are more effective than older ones in financial sentiment analysis, though the system carries significant risks when it comes to using synthetic or unverified textual input. Generative architectures also possess many obvious benefits over econometric models including adaptive learning, high-dimensional data synthesis, and modeling. Kumar *et al.* (2025) discover that the hybrid transformer-GAN makes predictions with 22% higher accuracy than ARIMA when predicting the qualitative and quantitative information. Sai *et al.* (2025) also demonstrate that a recent subdivision of the generative AI is diffusion models that are more reliable and realistic in providing synthetic data to market simulation, and thus increase the precision of stress testing and predictions of rare events. Transparency and interpretability are also important aspects of the old models of the economy, which are becoming less predictive and less conscious of the context and predictive ability of AI methods (Bartakovach *et al*., 2025). The growth of the generative finance is an area that has been a subject of study in a number of studies. Those are used in market forecasting and risk assessment (Hossain, 2025; Bai *et al*., 2024) as well as in business valuation and strategy modeling (Kanbach *et al*., 2024; Doshi *et al*., 2025). According to Saivasan and Lokhande (2023), a new feature refers to GANs in combination with the visualization of the metaverse, where 3D visual forms are used to emphasize the fact that the financial forecast made in a spreadsheet format is not very suitable and immersive 3D environment can significantly help in the real-time modelling of a situation and interactive decision-making. Equally, Sriram and Seenu (2023) demonstrate that repeated generative networks in integrated payment systems lessen transaction errors and improve ability to detect abnormalities, although the logic of their probabilistic nature is impractical to audit in compliance-intensive settings. However, despite the above performance advantages, generative models continue to have unaddressed interpretability, auditability and governance challenges. Krause (2023) and Piacentino (2025) emphasize the role of transformer-based systems as a kind of black boxes with which regulators and auditors do not know how to predict the outputs of the activities. Opposing this, de Kok (2025) expresses his concern about accounting research, where he observes Basing financial judgments on ChatGPT-based explanations may be wrongly tested on domain-specific fine-tuning. On the same note, Bartakova *et al.* (2025) emphasize that despite the fact that the generative system helps in the decision-making process, the opaque nature of the system lowers the financial accountability particularly when the artificial valuations produced by AI impact the capital markets or regulatory reports. Put collectively, these data mark that there is a danger of elimination of the institutional trust under the posing of the accuracy that cannot be clarified a paradox not unique to the conventional transparent econometric regimes. Regarding model risks and regulative accountability, Sourav *et al.* (2025) may suggest that generative systems may raise accumulative errors in the macroeconomic data at country scale through training on biased or partial macroeconomic data. Omoruyi (2025) continues to warn that reinforcement learning agents might be able to make adaptive judgements, however, they also might cause overfitting to historical volatility, leading to the policy outcomes being fragile or unstable. These remarks shed light on the fact that human monitoring is essential, particularly when AI environments are highly performing (Piacentino, 2025; Krause, 2023).

However, generative AI is a paradigmatic change in epistemology and a buildup in technology in the production and confirmation of financial judgement. Conventionally econometric models incorporate the notion of stability and linear causality through constant, unemodulated functional forms whereas generative models incorporate changing, dynamic, priors, which are learnt based on real-time feedback (Kanbach *et al.*, 2024). This is flexibility of enabling opportunities of designing business models, which gives organisations possibilities to experiment hypothetical values, market trends / investment or movements previously unfeasible due to the deterministic modeling.

5.2 Decision-making effects: How AI-driven analytics influence strategic decision-making in corporate finance and investment, and what evidence exists for improvements in accuracy, speed and cost efficiency

The key advantage of AI-driven analytics is that the predictive accuracy increases, Celestin and Mishra (2025) demonstrate that the production of GANs in the risk modeling methodologies is 24% more dependable than a customary model, and they concur with Kumar *et al.* (2025), that the term their importance to the application of AI is significant since it can bring to light tiny nonlinear patterns that are not present in their traditional model. Analytics is more effective with transformers than with ARIMA and VAR models, even in the volatile capital flow analysis, Hossain (2025) and Nweke and Adelusi (2025) also show. Che *et al.* (2024) also supplement these results by showing that implementing generative AI to predict financial markets enhances the time sensitivity of the model but also stabilizes the decision results when the data in these experience high frequency volatility. More importantly, yet, the reliance of the advances in accuracy on the quality of the training data introduces a structural vulnerability, as data bias can deprive learning results of mispricing the system or of misinformed investment expectations. According to Bartakova *et al.* (2025) AI systems excel in modeling accuracy, although they are affected by the context of their performance in terms of training representativeness and data openness, and high accuracy does not mean absolute reliability.

AI is a significant factor in the speed and agility of decision-making, and transformer-based analytics decreases data-to-decision time by 35 percent, according to Pillai (2023); and deep network automation speeds up decision time by 25 percent, according to Chowdhury (2024). This trend falls in tandem with the arguments made by Ravichandran *et al.*, (2024) that the generative AI automates business intelligence processes to the extent it allows decision-making and scenario simulations to be conducted at any given time. These swifter innovation can be applied to corporate finance with the aim of managing the risk and exploiting arbets by reacting to market volatility. Nevertheless, the option of shortening the time frame of deliberative supervision may be also effective through the acceleration of decision-making processes. Doshi *et al.* (2025) argue that speed brought about by AI in cases where executives make decisions without interpretation questions generates the speed/ slowness paradox, where the reaction time is slower than otherwise, and consequently, this can be used to exceed reflective judgement. Due to this, AI makes the decision-making process faster, yet it could also reduce the level of thought in making important financial decisions. According to Ahirrao *et al.* (2025) and Badmus *et al.* (2024), automated multi-scenario assessments and optimisation reduce the amount of repetitive analytical tasks and allow distributing financial resources more effectively. Saivasan and Lokhande (2023) also indicate that the automation explained by AI in metaverse-connected finance system reduces data processing costs by 18-22 percent and operational productivity through smart process integration. The same researchers (Ravichandran Sr. *et al.*, 2024) further substantiate this finding by demonstrating that with the help of AI-driven business intelligence automation human redundancy of pipelines of data can be reduced to a minimum and even lead to the development of cost saving that can be demonstrable without the compromising of any analytical richness. Those economic gains though leave out part of the intricate form of government. This deprivation of the human ability to interpret will confuse the chain of reasoning in decision making and this is particularly a nightmare to the financial auditors and regulators. The cost of implementing the mechanisms of the explanations and the lack of transparency should then be considered in comparison to the savings.

Besides performance, AI-based analytics are also connected with how organisations think, act and design decisions. According to Ghosh (2025), this is the emergence of algorithmic strategic intelligence that is the idea that AI systems integrate the constantly changing flows of data to build up the strategy together. Goel and Mahur (2025) further support this idea by claiming that AI-driven business analytics move decision-making towards the active modeling, and not the retrospective analysis. It is under this change that cross-functional consistency could easily be introduced as evidenced by Nweke *et al.*, (2025) when the integrations of predictive and prescriptive integrations of real-time finance, operations, and executives are introduced to present a unitary framework of decisions. Nevertheless, this coherence can also be an issue of dependency algorithm. As suggested by Chauhan *et al.* (2025), such excessive dependency on algorithmic thought can destroy managerial intuition and ethical thought especially in opaque black box models. AI offers greater level of strategic literacy, but it needs interpretative literacy of decision-makers to prevent the issue of technological determinism. The issues that arise regarding the utilization of AI-powered analytics are governance challenges related to the areas of accountability, prejudice, and transparency. Even the latest generative AI systems, including ChatGPT and Bard, are said to be hegemonic between the financial discourse and lack clear reasoning choices (Krause, 2023). This traceability is affected by such lack of transparency in the financial decision audit, which is a heavy burden in terms of regulatory compliance in corporate finance. According to Bartakova *et al.* (2025) and Badmus *et al.* (2024), the additional requests of explainable AI models that can incorporate both the performance and interpretability have been suggested; these offer human intervention in the flow of information automated decision-making. The issue of ethics is among the problems, which are essential as answering the question whether the strategic inclusion of AI makes it possible to keep on improving the quality of life or make the dependency even more severe. According to Sriram (2022), the implementation of the generative AI to the financial reporting entails open human-AI partnership to establish accountability and trust in the financial intelligence systems.

5.3 Human-AI integration and risks: Roles human expertise and oversight play in implementing generative AI for financial decisions, and what ethical, reliability and regulatory challenges have been identified

One of the most significant concerns in the integration of financial AI is a loss of human intuition as generative systems become more independent. Holzinger *et al.* (2025) believe that human monitoring of high-frequency trading and automated underwriting has become entirely symbolic, because algorithmic operations are extraordinarily quick and complicated and far surpass human comprehension levels. This opinion is further supported by Bartáková *et al.* (2025), who point out that while generative models are a great way to combine various financial datasets for forecasting, their decision-making is still mostly opaque and requires interpretive validation after the fact to guarantee reliability and regulatory compliance. The gap in interpretability is also noted by Krause (2023), who explored large language models like ChatGPT and Bard in financial analytics. These models, Krause argues are black boxes that, though analytically admirable, produce outcomes that are contextually persuasive, but cognitively imperceptible. On the same note, Sriram (2022) asserts that implementing generative AI in financial reporting systems enhances the effectiveness of analytical operations at the expense of the transparency of causal reasoning.

Ethical responsibility grows as financial institutions develop and grow more dependent on generative AI. In Raza *et al.* (2025), the essence of mistakes in AI-financial systems is often not clearly defined because of flawed data, biased training data, or architectural mistakes that make it difficult to identify the causes of the fault. Under such conditions, human control develops into reactive control rather than active ethical design and control. As Chauhan *et al.* (2025) note, human expertise is now a kind of a safeguard of last resort, particularly in high-stakes domains of decisions, e.g., the risk assessment and investment strategy. Human beings act as a liaison between algorithms and situations, and it is intended to lessen the automation bias by means of moral judgment. Goel and Mahur (2025) also propose that financial professionals are supposed to act as AI-enhanced interpreters who can appropriately and ethically contextualize AI outputs. However, these expectations are based on the presumption of an AI knowledge base that most businesses do not have a gap that risks transforming human control into formal adherence as opposed to actual moral engagement.

The financial regulation of AI is chaotic and unpredictable. Both Tekale (2024) and G'sell (2024) indicate that despite the presence of new policy frameworks where bias detection, explainability, and fairness auditing are required, they are still in diverse ways enforced across countries. Regulatory asymmetry was specifically cited in a study published in Saha *et al.*, (2025) as the primary factor that banks should raise concerns about to prevent the use of AI regarding cross-border data control and the obligation of algorithms as a regulatory factor, which is unpredictable. To substantiate this, Saivasan and Lokhande (2023) assert that the fast development of the generative AI and metaverse-linked financial systems is a failure of the current policy structures where the enterprises are left at the stage of making compliance judgements in ambiguity. G'sell (2024) presents a good argument in favour of risk-based adaptive regulation, since there should always be a balance between technology innovation and governance as a part of the governance structure. In the absence of this, generative AI will become a regulatory blind spot, which is enhanced by structural vulnerabilities in capital markets and consumer finance.

Aldasoro *et al.* (2024) and Chauhan *et al.* (2025) both arrive at a conclusion about the fact that human competence is not outdated, though it is redefined in the age of AI-driven finance. Aldasoro refers to this as AI risk literacy and refers to it as now supervision as necessitating a combination of quantitative literacy, ethical reasoning, and interpretive ability. This mixed power will not just allow humans to become fail-safes, but co-rule and moral overseers of artificial intelligence. Ravichandran *et al.* (2024) further demonstrated that contextual correction, where individuals are made aware of model anomalies and change outputs with domain experience, improves the decision quality in cases where human analysts are introduced as part of AI-based loops in business intelligence. Sriram and Seenu (2023) also stress the necessity of human in command in automated payment and trading systems, so that ethical check and responsibility can be embedded within financial automation systems.

The bias, explanation, and agency attribution spreading are the most salient ethical issues. Hacker *et al*., (2023) provide a wide range of regulatory definitions, which involve pre-deployment audits, constant interpretation test, and post-deployment monitoring. According to Bartakova *et al.* (2025), explainability interfaces can be used to enable auditors and compliance officers to pursue newly-caused pathways in AI-based financial outputs. In the absence of transparency scaffolding, generative systems have a danger of perpetuating an algorithmic obscurity, in which the conclusion has been reached as true mathematically, yet is either unethical or culturally objectionable. The financial industry has entered a new stage of morality, just as it is planned by Kalia (2023): the institutions will have to define who bears the final responsibility in case of a financial misjudgment, as the generative AI is gaining more cognitive powers. The need to assert legitimacy and culpability in AI-mediated monetary ecosystems, which this ethical issue highlights, is constant.

### Recommendations for implementation

In order to facilitate the adequate and efficient use of generative AI in the financial decisions process, organisations have to create the model of balance, which is governance-based and considers human experience, ethical principles, and regulatory discretion. First, the companies must establish cross-functional oversight boards to have a combination of data scientists, financial analysts, ethicists and compliance officers to ensure that the AI models are technically streamlined yet socially and ethically oriented. The human monitoring at each and every phase of the AI life cycle collection, modeling, deployment, and post decision auditing will ensure that accountability remains constant and data gaps are not exacerbated. On the same note, organisations are advised to look into their own model design and repeated verification as explainability measures. Installing interpretive interfaces to translate the algorithmic reasoning into rational inferences enables human experts to discuss, corroborate, as well as build the knowledge that AI retrieves. In addition, the corporations ought to invest in AI literacy courses to provide a more thorough training to workers to reason ethically and possess technological skills such that human participants play an active role in regulation rather than passively observe.

Besides that, risk management should become more ethical by introducing fairness and bias detectors as well as ethical checks in the AI development pipeline to ensure that compliance is not a mandatory requirement but a proactive security control measure instead. Co-operation with the regulators, industry players and civil society organisation can also enhance transparency and harmonisation of the global standards; reducing regulatory asymmetry between the markets. Finally, AI should be used as an enhancer and not a decision maker through cultivating a culture of human-in-command responsibility by financial institutions. This provides a balance between the ability of human judgment to make a decision in a tactical financial context with the precision of the AI analysis. As part of technology and human responsibility, institutions can embrace the revolutionary potential of generative AI, and still uphold trust, transparency, and ethical integrity that forms the foundation of stable financial systems.

## CONCLUSION

This systematic review examined the overlaps between generative artificial intelligence (GAI) and latest analytics and financial strategic decision-making. This review pointed that GAI is a paradigm shift to the old econometric and predictive analytic systems to adaptive and data-driven and cognitively enhanced financial intelligence systems. Nonlinear pattern recognition, contextual synthesis, and scenario simulation have altered financial modelling, valuation and forecasting by using GANs, transformer-based LLMs, and diffusion architectures instead of the traditional econometric models. Also, in this study, GAI was found to be more precise, quicker and less expensive than previous techniques in corporate finance and investment decisions. enhancing accuracy of forecasting and quicker decision making processes. However, there are political and ethical side effects that are associated with these performance gains. From this study, explainability, bias, and accountability are the challenges of institutional adoption of GAI, along with less human ambiguity or decision logic. Human oversight becomes therefore a basic one but not the passive one but rather the active co-regulatory tool that guarantees the ethical compliance, contextual interpretation and regulatory compliance. This study supports the importance of human expertise to be considered a part of the AI lifecycle so that high-stakes financial markets can maintain accountability and trust.

Lastly, the human judgement and the balance of algorithmic intelligence remains important in financial modelling and strategic decision making. Organizations that share ideals of leadership which focus on transparency, fairness and accountable human-in-commands are more likely to exploit the force of GAI to transform and mitigate its systemic and moral threats. The generative AI is not just a technology, but rather, this is a way of reinventing the epistemology, ethics, and strategic reasoning of the present-day finance.

## REFERENCES

Ahirrao, Y. S., Ansari, I., Azim, K. S., Bhujel, K., & Panchal, S. S. (2025). AI-Powered Financial Strategy: Transforming Business Decision-Making Through Predictive Analytics. *Emerging Frontiers Library for The American Journal of Engineering and Technology, 7*(09), 126-151.

Aldasoro, I., Gambacorta, L., Korinek, A., Shreeti, V., & Stein, M. (2024). Intelligent financial system: how AI is transforming finance.

Badmus, O., Rajput, S., Arogundade, J., & Williams, M. (2024). AI-driven business analytics and decision

making. *World Journal of Advanced Research and Reviews, 24*(1), 616-633.

Bai, X., Zhuang, S., Xie, H., & Guo, L. (2024). Leveraging generative artificial intelligence for financial market trading data management and prediction.

Bartáková, G. P., Almadhor, A., Qayyum, A., Abeer, K., & Durrani, A. (2025). Evaluating the capacity and limitations of generative AI in financial decision making. *Computer Standards & Interfaces, 93,* 103965.

Celestin, M., & Mishra, A. K. (2025). AI-driven financial analytics: Enhancing forecast accuracy, risk management, and decision-making in corporate finance. *Janajyoti Journal, 3*(1), 1-27.

Chauhan, N., Thakur, G., Joshi, A., Kumar, V., Kumar, A., & Singh, Y. (2025). The Transformative Impact of AI Technologies on Decision-Making Processes and Operational Efficiency Across Sectors, with a Focus on Finance. In Generative AI in FinTech: Revolutionizing Finance Through Intelligent Algorithms (pp. 73-113). Cham: Springer Nature Switzerland.

Che, C., Huang, Z., Li, C., Zheng, H., & Tian, X. (2024). Integrating generative ai into financial market prediction for improved decision making. arXiv preprint arXiv:2404.03523.

Chowdhury, R. H. (2024). AI-driven business analytics for operational efficiency. *World Journal of Advanced Engineering Technology and Sciences, 12*(2), 535-543.

de Kok, T. (2025). ChatGPT for textual analysis? How to use generative LLMs in accounting research. Management Science.

Doshi, A. R., Bell, J. J., Mirzayev, E., & Vanneste, B. S. (2025). Generative artificial intelligence and evaluating strategic decisions. *Strategic Management Journal, 46*(3), 583-610.

Garcia, A., & Adams, J. (2022). Data-Driven decision making: leveraging analytics and AI for strategic advantage. *Research Studies of Business, 1*(02), 77-85.

Ghosh, U. K. (2025). Transformative AI Applications in Business Decision-Making: Advancing Data-Driven Strategies and Organizational Intelligence. In AI-Powered Leadership: Transforming Organizations in the Digital Age (pp. 1-40). IGI Global Scientific Publishing.

Goel, P. K., & Mahur, L. S. (2025). Impact of Generative AI on Business Analytics and Decision Making in Service Organizations. In Generative AI for Business Analytics and Strategic Decision Making in Service Industry (pp. 65-88). IGI Global Scientific Publishing.

G'sell, F. (2024). Regulating under uncertainty: Governance options for generative AI. Available at SSRN 4918704.

Hacker, P., Engel, A., & Mauer, M. (2023, June). Regulating ChatGPT and other large generative AI models. In Proceedings of the 2023 ACM conference on fairness, accountability, and transparency (pp. 1112-1123).

Holzinger, A., Zatloukal, K., & Müller, H. (2025). Is human oversight to AI systems still possible?. *New Biotechnology, 85*, 59-62.

Hossain, M. A. (2025). Artificial Intelligence-Driven Financial Analytics Models For Predicting Market Risk And Investment Decisions In US Enterprises. ASRC Procedia: *Global Perspectives in Science and Scholarship, 1*(01), 1066-1095.

Jowarder, M. I. J. R. A. (2024). AI-Driven Strategic Insights: Enhancing Decision-Making Processes in Business Development.

Kalia, S. (2023). Potential impact of generative artificial intelligence (AI) on the financial industry. *International Journal on Cybernetics & Informatics (IJCI), 12*(12), 37.

Kanbach, D. K., Heiduk, L., Blueher, G., Schreiter, M., & Lahmann, A. (2024). The GenAI is out of the bottle: generative artificial intelligence from a business model innovation perspective. *Review of Managerial Science, 18*(4), 1189-1220.

Krause, D. (2023). Large language models and generative AI in finance: an analysis of ChatGPT, Bard, and Bing AI. Bard, and Bing AI (July 15, 2023).

Kumar, T., Lalar, S., Garg, V., Sharma, P., & Mishra, R. D. (2025). Generative Artificial Intelligence (GAI) for Accurate Financial Forecasting. Generative Artificial Intelligence in Finance: Large Language Models, Interfaces, and Industry Use Cases to Transform Accounting and Finance Processes, 57-76.

Kumari, P., Singh, S. K., & Utpal, V. K. J. (2025). Guardians of Accountability: The Role of Media in Oversight and Governance of Generative AI Applications in Fintech. In Generative AI in FinTech: Revolutionizing Finance Through Intelligent Algorithms (pp. 345-359). Cham: Springer Nature Switzerland.

Nweke, O., & Adelusi, O. (2025). Utilizing AI driven forecasting, optimization, and data insights to strengthen corporate strategic planning. *International Journal of Research Publication and Reviews, 6*(3), 4260-4272.

Nweke, O., & Owusu-Berko, L. (2025) Integrating AI-driven predictive and prescriptive analytics for enhancing strategic decision-making and operational efficiency across industries.

Omoruyi, N. (2025). Advanced Computational Methods for Financial Planning and Analysis Risk Assessment using Data Science-Driven Model Validation Techniques. *International Journal of Research Publication and Reviews.* doi: https://www. semanticscholar. org/er/4d27d96c40e20c0bd7df2d9220bd4b355a381c82.

Page, M. J., McKenzie, J. E., Bossuyt, P. M., Boutron, I., Hoffmann, T. C., Mulrow, C. D., ... & Moher, D. (2021). The PRISMA 2020 statement: an updated guideline for reporting systematic reviews. bmj, 372.

Pattanayak, S. K. (2022). Generative AI for market analysis in business consulting: Revolutionizing data insights and competitive intelligence. *International Journal of Enhanced Research in Management & Computer Applications, 11,* 74-86.

Piacentino, L. (2025). The Transformative Role of Artificial Intelligence in Financial Decision-Making: Main Applications in Corporate and Personal Finance, Impacts and Future Prospects (Doctoral dissertation, Politecnico di Torino).

Pillai, V. (2023). Integrating ai-driven techniques in big data analytics: Enhancing decision-making in financial markets. *International Journal of Engineering and Computer Science, 12*(07), 10-18535.

Ravichandran Sr, P., Machireddy Sr, J. R., & Rachakatla, S. K. (2024). Harnessing Generative AI for Automated Data Analytics in Business Intelligence and Decision-Making. *Hong Kong Journal of AI and Medicine, 4*(1), 122-145.

Ravichandran, P., Machireddy, J. R., & Rachakatla, S. K. (2024). Generative AI in Business Analytics: Creating Predictive Models from Unstructured Data. *Hong Kong Journal of AI and Medicine, 4*(1), 146-169.

Raza, S., Qureshi, R., Zahid, A., Fioresi, J., Sadak, F., Saeed, M., ... & Shoman, M. (2025). Who is responsible? the data, models, users or regulations? responsible generative ai for a sustainable future. Authorea Preprints.

Saha, B., Rani, N., & Shukla, S. K. (2025). Generative AI in Financial Institution: A Global Survey of Opportunities, Threats, and Regulation. arXiv preprint arXiv:2504.21574.

Sai, S., Arunakar, K., Chamola, V., Hussain, A., Bisht, P., & Kumar, S. (2025). Generative AI for finance: applications, case studies and challenges. *Expert Systems, 42*(3), e70018.

Saivasan, R., & Lokhande, M. (2023). Exploring use cases of generative AI and metaverse in financial analytics: Unveiling the synergies of advanced technologies. *International Journal of Global Business and Competitiveness, 18*(Suppl 1), 77-86.

Sharma, P. (2023). Analyzing How Rigorous Financial Analysis Informs Strategic Decisions and Contributes to Corporate Growth. *Nanotechnology Perceptions, 20,* 219-229.

Sourav, M. S. A., Asha, N. B., & Reza, J. (2025). Generative AI in Business Analytics: Opportunities and Risks for National Economic Growth. *Journal of Computer Science and Technology Studies, 7*(11), 224-247.

Sriram, H. K. (2022). Integrating generative AI into financial reporting systems for automated insights and decision support. Available at SSRN 5232395.

Sriram, H. K., & Seenu, A. (2023). Generative AI-Driven Automation in Integrated Payment Solutions: Transforming Financial Transactions with Neural Network-Enabled Insights. *International Journal of Finance (IJFIN), 36*(6), 70-95.

Tekale, K. M. (2024). AI Governance in Underwriting and Claims: Responding to 2024 Regulations on Generative AI, Bias Detection, and Explainability in Insurance Decisioning. *International Journal of AI, BigData, Computational and Management Studies, 5*(1), 159-166.